\documentclass[runningheads]{llncs}
\usepackage[T1]{fontenc}
\usepackage{graphicx}
\usepackage[normalem]{ulem}
\usepackage{fdl}

\usepackage[capitalize]{cleveref}
\crefname{section}{Sec.}{Secs.}
\Crefname{section}{Section}{Sections}
\Crefname{table}{Table}{Tables}
\crefname{table}{Tab.}{Tabs.}

\begin{document}

\title{Adaptive Supervised PatchNCE Loss for Learning H\&E-to-IHC Stain Translation with Inconsistent Groundtruth Image Pairs}

\titlerunning{Adaptive Supervised PatchNCE}
%

\author{Fangda Li\textsuperscript{1*} \and Zhiqiang Hu\textsuperscript{2} \and Wen Chen\textsuperscript{2} \and Avinash Kak\textsuperscript{1}}


\authorrunning{Li et al.}

\institute{Electrical and Computer Engineering, Purdue University, USA \\
\email{\{li1208,kak\}@purdue.edu} \\
\and
Sensetime Research, China \\
\email{\{huzhiqiang,chenwen\}@sensetime.com} \\
*Corresponding author
}

\maketitle              

\begin{abstract}
Immunohistochemical (IHC) staining highlights the molecular information critical to diagnostics in tissue samples.
However, compared to H\&E staining, IHC staining can be much more expensive in terms of both labor and the laboratory equipment required.
This motivates recent research that demonstrates that the correlations between the morphological information present in the H\&E-stained slides and the molecular information in the IHC-stained slides can be used for H\&E-to-IHC stain translation.
However, due to a lack of pixel-perfect H\&E-IHC groundtruth pairs, most existing methods have resorted to relying on expert annotations.
To remedy this situation, we present a new loss function, \textit{Adaptive Supervised PatchNCE} (ASP), to directly deal with the input to target inconsistencies in a proposed H\&E-to-IHC image-to-image translation framework.
The ASP loss is built upon a patch-based contrastive learning criterion, named \textit{Supervised PatchNCE} (SP), and augments it further with weight scheduling to mitigate the negative impact of noisy supervision.
Lastly, we introduce the \textit{Multi-IHC Stain Translation} (MIST) dataset, which contains aligned H\&E-IHC patches for 4 different IHC stains critical to breast cancer diagnosis.
In our experiment, we demonstrate that our proposed method outperforms existing image-to-image translation methods for stain translation to multiple IHC stains.
All of our code and datasets are available at \texttt{\url{https://github.com/lifangda01/AdaptiveSupervisedPatchNCE}}.


\keywords{Generative Adversarial Network \and Contrastive Learning \and H\&E-to-IHC Stain Translation.}
\end{abstract}

\section{Introduction}
Immunohistochemical (IHC) staining is a widely used technique in pathology for visualizing abnormal cells that are often found in tumors.
IHC chromogens highlight the presence of certain antigens or proteins by staining their corresponding antibodies.
For instance, the HER2 (human epidermal growth factor receptor 2) biomarker is associated with aggressive breast tumor development and is essential in forming a precise treatment plan. 
Despite its capability to provide highly valuable diagnostic information, the process of IHC staining is very labor-intensive, time-consuming and requires specialized histotechnologists and laboratory equipments \cite{anglade2020can}.
Such factors hinder the general availability of IHC staining in histopathological applications.

At the other end of the spectrum, H\&E (Hematoxylin and Eosin) staining, as the gold standard in histological staining, highlights the tissue structures and cell morphology. 
In routine diagnostics, on account of its much lower cost, an H\&E-stained slide is prepared by pathologists in order to determine whether or not to also apply the IHC stains for a more precise assessment of the disease. 
Therefore, it is of great interest to have an algorithm that can automatically translate an H\&E-stained slide into one that could be considered to have been stained with IHC while accurately predicting the target expression levels.

To that end, researchers have recently proposed to use GAN-based Image-to-Image Translation (I2IT) algorithms for transforming H\&E-stained slides into IHC.
Despite the progress, the outstanding challenge in training such I2IT frameworks is the lack of aligned H\&E-IHC image pairs, or in other words, the inconsistencies in the H\&E-IHC groundtruth pairs.
To explain, since re-staining a slice is physically infeasible, a matching pair of H\&E-IHC slices are taken from two depth-wise \textit{consecutive cuts} of the same tissue then stained and scanned separately.
This inevitably prevents pixel-perfect image correspondences due to the slice-to-slice changes in cell morphology, staining-induced degradation (\eg tissue-tearing), imaging artifacts that may vary among slices (\eg camera out-of-focus) and multi-slice registration errors. 
An example pair of patches is shown in \cref{fig:spnce} and another pair with significant inconsistencies is shown in \cref{fig:similarites}(a)(c).
In the latter, comparing the groundtruth IHC image to the input H\&E image, one can clearly see the inconsistencies -- nearly the entire left half of the tissue present in the H\&E image is missing.

As a result, recent advances in H\&E-to-IHC I2IT have mostly avoided using the inconsistent GT pairs and instead have imposed the cycle-consistency constraint \cite{lin2022unpaired,liu2021unpaired,zeng2022semi}.
Moreover, existing approaches have also exploited using expert annotations such as per-cell labels \cite{liu2020predict}, semantic masks \cite{liu2021unpaired} and patch-level labels \cite{liu2021unpaired,zeng2022semi}.
As for the prior works that directly utilize the H\&E-IHC pairs for supervision, a variant of Pix2Pix \cite{isola2017image} that uses a Gaussian Pyramid based reconstruction loss to accommodate the noisy GT is proposed in \cite{liu2022bci}.
However, the robustness of such approaches that punish absolute errors in the generated image to dealing with GT inconsistencies remains unclear.

In this paper, we argue that the IHC slides, despite the disparities vis-a-vis their H\&E counterparts, can still serve as useful targets for stain translation. 
The work we present in this paper is based on the important realization that even when pairs of consecutive tissue slices do not yield images that are pixel-perfect aligned, it is highly likely that the corresponding patches in the two stains share the same diagnostic label.
For example, if the levels of expression in a region of the HER2 slide are high, the corresponding region in the H\&E slide is highly likely to contain a high density of cancerous cells.
Therefore, we set our goal to meaningfully leverage such correlations to benefit the H\&E-to-IHC I2IT while being resilient to any inconsistencies.

Toward this goal, we propose a supervised patchwise contrastive loss named the \textit{Adaptive Supervised PatchNCE} (ASP) loss.
Our formulation of this loss was inspired by the recent research findings that contrastive loss benefits model robustness under label noise \cite{xue2022investigating,ghosh2021contrastive}.
Furthermore, based on the observation that any dissimilarity between the patch embeddings at corresponding locations in the generated and groundtruth IHC images is indicative to the level of inconsistency of the GT at that location, we employ an adaptive weighting scheme in ASP.
By down-weighting the contrastive loss at locations with low similarities, \ie high inconsistencies, our proposed ASP loss helps the network learn more robustly.

Lastly, to support further research in virtual IHC-restaining, we present the \textit{Multi-IHC Stain Translation} (MIST) as a new public dataset.
The MIST dataset contains 4k+ training and 1k testing aligned H\&E-IHC patches for each of the following IHC stains that are critical for \textit{breast cancer} diagnostics: HER2, Ki67, ER (Estrogen Receptor) and PR (Progesterone Receptor).
We evaluated existing I2IT methods and ours for multiple IHC stains and demonstrate the superior performance achieved by our method both qualitatively and quantitatively.

\section{Method Description}
\subsection{The Supervised PatchNCE (SP) Loss}

\begin{figure}[t]
\centering
\includegraphics[width=0.9\textwidth]{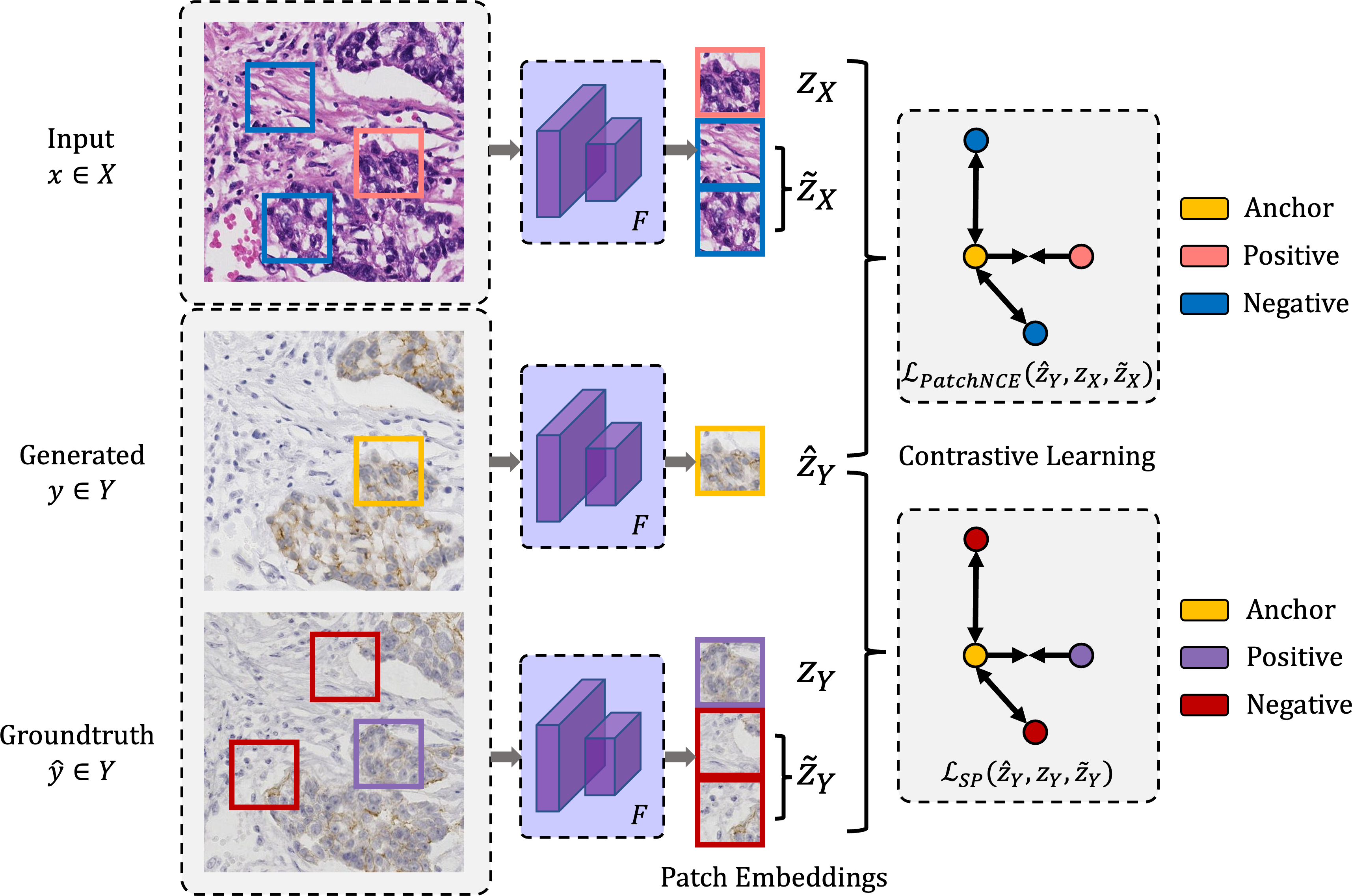}
\caption{
Illustration of the PatchNCE loss from \cite{park2020contrastive} and the Supervised PatchNCE (SP) loss.
The patch embeddings $\vz$ are extracted by a shared network $F$.
}
\label{fig:spnce}
\end{figure}

Before getting to our ASP loss, we need to first introduce the SP loss as a robust means to learning from inconsistent GT image pairs.
The SP loss was inspired by the findings in recent literature that demonstrate the positive effect of contrastive learning on boosting model robustness against label noise \cite{xue2022investigating,ghosh2021contrastive,zheltonozhskii2022contrast}. 
It takes the same form as the PatchNCE loss as introduced in \cite{park2020contrastive}, except that it is applied on the generated-GT image pair (instead of the input-generated pair).

The goal of the PatchNCE loss is to ensure the content is consistent across translation by maximizing the mutual information between the input and the corresponding output.
It does so by minimizing a patch-based InfoNCE loss \cite{oord2018representation}, which encourages the network to associate the corresponding patches with each other in the learned embedding space, while disassociating them from the noncorresponding ones. 
Mathematically, the InfoNCE loss takes the form:
\begin{equation}
\label{eq:infonce}
\calL_{\text{InfoNCE}}(\vv, \vv^+, \vv^-) = -\log \left[ \frac{\exp{(\vv \cdot \vv^+ / \tau)}}{\exp{(\vv \cdot \vv^+ / \tau)} + \sum_{n=1}^N \exp{(\vv \cdot \vv_n^- / \tau)}} \right],
\end{equation}
where $\vv$, $\vv^+$ and $\vv^-$ are the embeddings of the anchor, positive and negative samples, respectively.
With InfoNCE, the PatchNCE loss is set up as follows: given the anchor embedding $\vzhat_Y$ of a patch in the output image, the positive $\vz_X$ is the embedding of the corresponding patch from the input image, while the negatives $\vztilde_X$ are embeddings of the non-corresponding ones, \ie $\calL_{\text{PatchNCE}} = \calL_{\text{InfoNCE}}(\vzhat_Y, \vz_X, \vztilde_X)$.

As for the SP loss, given the embedding of an output patch $\vzhat_Y$ as anchor, we now designate the embedding of the corresponding patch in the groundtruth image $\vz_Y$ as the positive and the embeddings of the non-corresponding ones $\vztilde_Y$ as the negatives. 
We then use the same InfoNCE-based contrastive learning objective, \ie $\calL_{\text{SP}} = \calL_{\text{InfoNCE}}(\vzhat_Y, \vz_Y, \vztilde_Y)$. 
A depiction of both the PatchNCE loss and the SP loss is given in \cref{fig:spnce}.
It is worthy to note that, although a similar patchwise contrastive loss has been proposed in \cite{andonian2021contrastive} for supervised I2IT, it is one of our contributions in this paper to explicitly exploit the robustness of this contrastive loss in the context of H\&E-to-IHC translation where the GT pairs can be highly inconsistent for reasons mentioned previously.
We think that the key factor behind the robustness of $\calL_{\text{SP}}$ towards inconsistent GT compared to, say, the MSE loss, is its relativeness.
Instead of using an absolute loss term that may not work well on inconsistent groundtruth pairs, $\calL_{\text{SP}}$ punishes dissimilarities between the anchor and the positive in a learned latent space, relative to those between the anchor and the negatives. 

\subsection{The Adaptive Supervised PatchNCE (ASP) Loss}

\begin{figure}[t]
\centering
\includegraphics[width=0.9\textwidth]{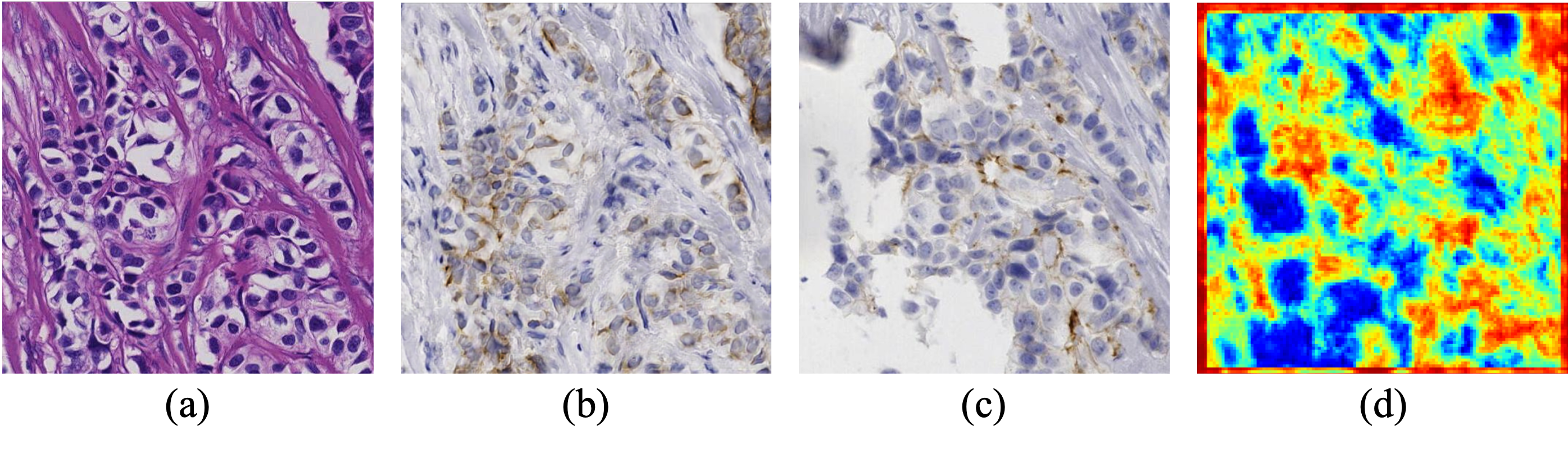}
\caption{
(a) Input H\&E image $\vx$, (b) generated IHC image $\vyhat$, (c) groundtruth IHC image $\vy$, and (d) heat map of the anchor-positive cosine similarities produced by a trained network at corresponding locations: $C_{s} = \vz_{\vyhat}^s \cdot \vz_{\vy}^s$, where $s$ is index of the spatial location.
}
\label{fig:similarites}
\end{figure}

To learn selectively from more consistent groundtruth locations, we further propose to augment the Supervised PatchNCE loss in an adaptive manner.
The key idea here is to automatically recognize patch locations that are inconsistent and adapt the SP loss so that the severely inconsistent patch locations will have lesser effects on training.
To measure the consistency at a given patch location, we use the cosine similarity between the embeddings of the generated IHC patch and the corresponding GT patch. 
In \cref{fig:similarites}, we show an example pair of generated vs GT IHC images that contain significant inconsistencies and their anchor-positive similarity heat map.
For pairs of embeddings produced by a \textit{trained} network, a high similarity value indicates good correspondence between the groundtruth patches while a low similarity value indicates inconsistencies.

\begin{figure}[t]
\centering
\includegraphics[width=0.9\textwidth]{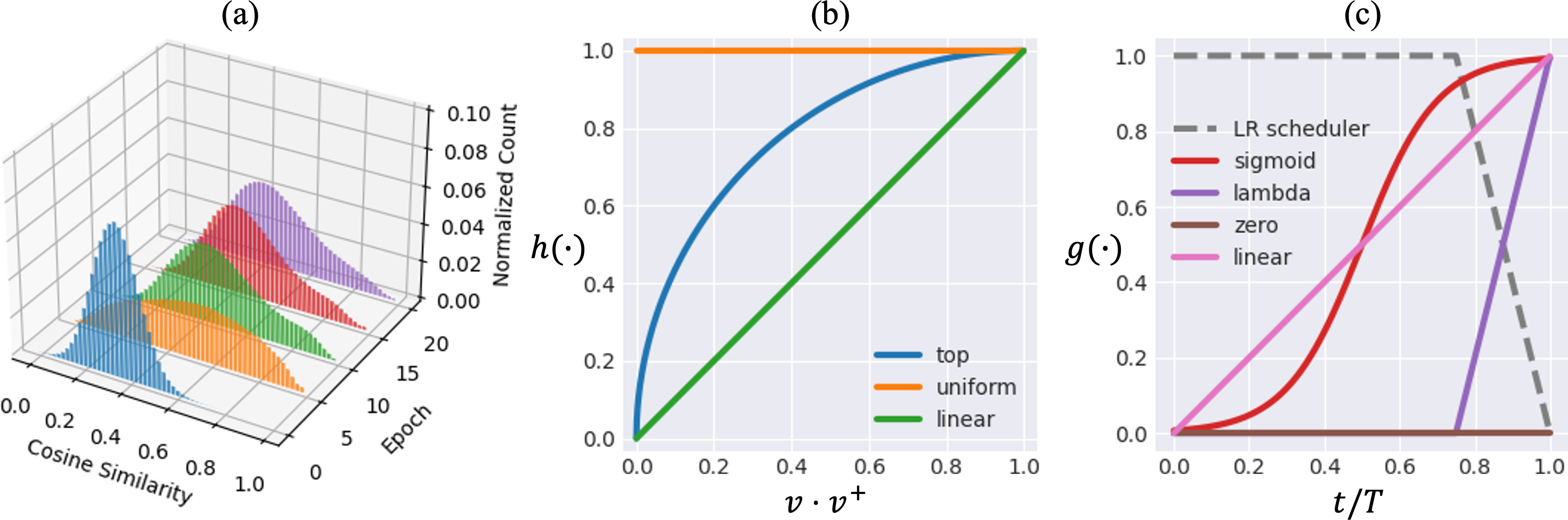}
\caption{
(a) Histograms of the anchor-positive similarity values at different epochs;
(b) The weight $h(\cdot)$ as a function of $C_{s}$;
(c) The scheduling function $g(\cdot)$.}
\label{fig:functions}
\end{figure}

Directly motivated by this observation, we first propose a weighting scheme for the SP loss.
More specifically, we assign lower weights to patch locations that have low anchor-positive similarity values to alleviate the negative impacts the inconsistent targets may have on training.
At training time $t$, the weight is a function of the anchor-positive cosine similarity.
Examples of the weight function $h(\cdot)$ are shown in \cref{fig:functions}(b). 
The weight functions are monotonic increasing so that the more confident patch locations are always treated with more importance.

In order to make the weighting scheme work in practice, we must also account for the phase of training.
The intuition is that, during the initial phase of training, the network is not going to be able to discriminate between consistent patch locations from those that are inconsistent. 
Additionally, as shown in \cref{fig:functions}(a), the histograms of the anchor-positive similarity evolve rather slowly over the training epochs.
Therefore, it would not make sense to reinforce the weighting function in the beginning of the training as much as near the end of the training.

\begin{figure}[t]
\centering
\includegraphics[width=0.99\textwidth]{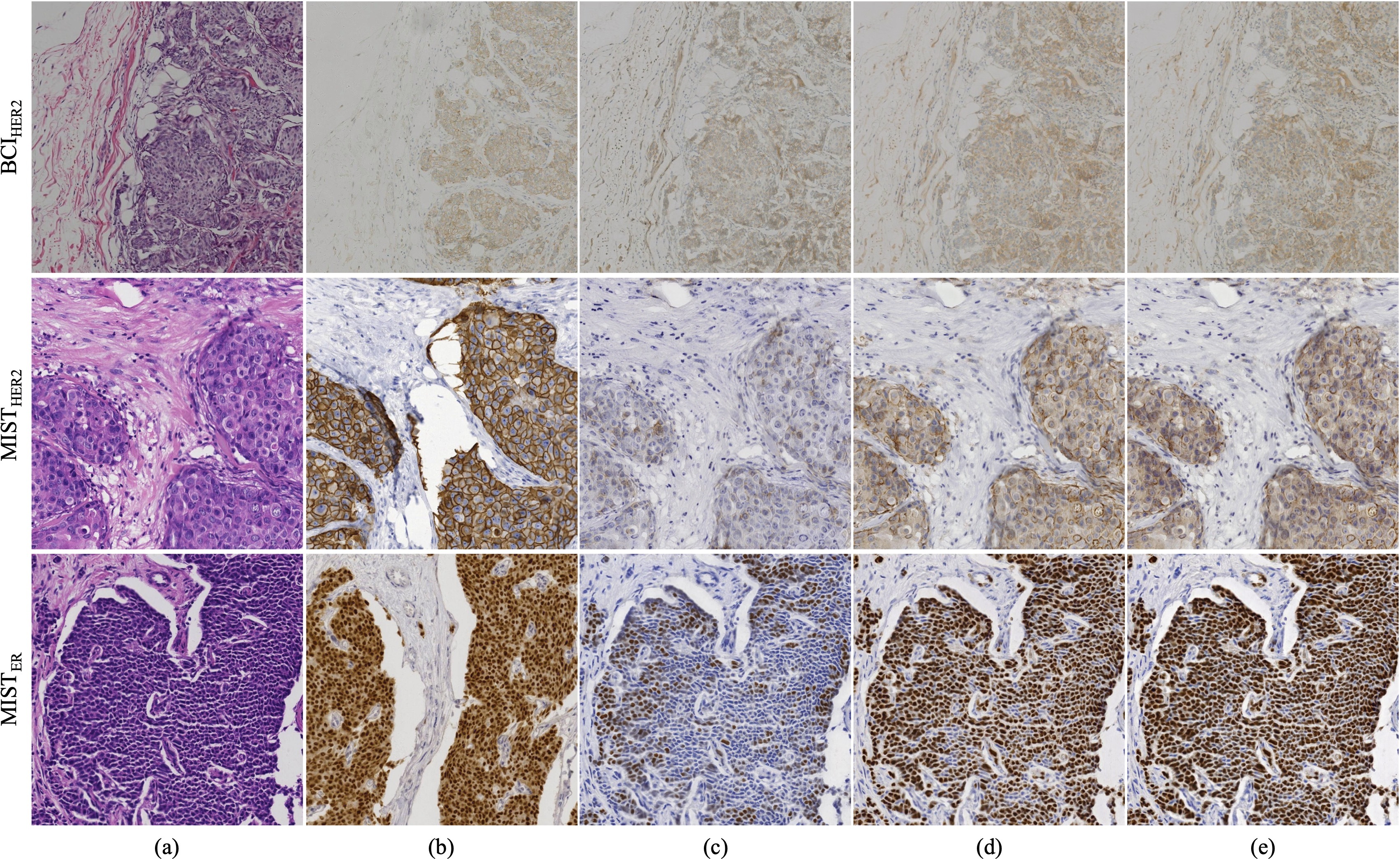}
\caption{
Left to right: (a) Input H\&E image; (b) Groundtruth IHC image; (c) Generated image without $\calL_{\text{SP}}$; (d) With $\calL_{\text{SP}}$; (e) With $\calL_{\text{ASP}}^{\text{(lambda,linear)}}$.}
\label{fig:qualitative}
\end{figure}

To that end, we further augment the weight so that it is also a function of the training iterations. 
Such scheduling of the weights is done so that in the beginning of the training, the weights are uniform in order not to wrongly bias the network when the embeddings are still indiscriminative. 
And as training progresses, the selective weighting scheme is gradually enforced so that the inconsistent patch locations are treated with reduced weights.
We call this gradual process of shifting the learning focus \textit{weight scheduling}.
Let $t$ denote the current iteration and $T$ the total number of training iterations.
Then weight scheduling is achieved by using a scheduling function $g(\frac{t}{T})$.
Various options of $g(\cdot)$ are shown in \cref{fig:functions}(c).
Subsequently, combining the weighting function with the scheduling function, we can write the following formula for the final weight:
\begin{equation}
w_t (\vv, \vv^+) = \left(1 - g\left(\frac{t}{T}\right)\right) \times 1.0 + g\left(\frac{t}{T}\right) \times h(\vv \cdot \vv^+).
\end{equation}
We refer to the new augmented Supervised PatchNCE loss as the Adaptive Supervised PatchNCE (ASP) loss, which can be expressed as:
\begin{equation}
\calL_{\text{ASP}}(G, H, X, Y, t) = \E_{(\vx, \vy) \sim (X, Y)} \sum_{l=1}^L \sum_{s=1}^{S_l} \frac{w_t(\vzhat^{l, s}_Y, \vz^{l, s}_Y)}{W_{t}^l} \cdot \calL_{\text{InfoNCE}}(\vzhat^{l, s}_Y, \vz^{l, s}_Y, \vztilde^{l, s}_Y),
\end{equation}
where $W_{t}^l = \sum_s w_t^{l, s}$ is a normalization factor that maintains the total magnitude of the loss after applying the weights.
Finally, the overall learning objective for our generator is as follows:
\begin{equation}
\begin{aligned}
\calL_{\text{adv}} + \lambda_{\text{PatchNCE}} \calL_{\text{PatchNCE}} + \lambda_{\text{ASP}} \calL_{\text{ASP}} + \lambda_{\text{GP}} \calL_{\text{GP}},
\end{aligned}
\label{eq:overall}
\end{equation}
where $\calL_{\text{GP}}$ is the Gaussian Pyramid based reconstruction loss from \cite{liu2022bci}.

\section{Experiments}

\begin{table}[t]
\caption{
Quantitative evaluations on all three datasets using both paired and unpaired metrics. 
The \textbf{best} values are highlighted. 
Ours is with $\calL_{\text{ASP}}^{\text{(lambda,linear)}}$.
}
\label{tbl:competing}
\centering
\begin{tabular}{clcccccccc}
\hline
\multirow{2}{*}{Dataset} & \multicolumn{1}{c}{\multirow{2}{*}{Method}} & \multirow{2}{*}{SSIM$\uparrow$} & \multicolumn{5}{c}{$\text{PHV}_{T=0.01}$$\downarrow$} & \multirow{2}{*}{FID$\downarrow$} & \multirow{2}{*}{KID$\downarrow$} \\
 & \multicolumn{1}{c}{} &  & layer1 & layer2 & layer3 & layer4 & avg. &  &  \\ \hline
 \multirow{5}{*}{$\text{BCI}_\text{HER2}$} & CycleGAN & 0.4424 & 0.4264 & 0.3924 & 0.2610 & 0.7262 & 0.4515 & \textbf{63.5} & 10.7 \\
 & CUT+$\calL_{\text{GP}}$ & 0.4802 & \textbf{0.4263} & 0.3784 & 0.2364 & 0.7328 & 0.4435 & 65.0 & 10.9 \\
 & Pix2Pix & 0.4372 & 0.5121 & 0.4531 & 0.2953 & 0.7484 & 0.5022 & 100.0 & 44.6 \\
 & PyramidP2P & 0.5001 & 0.4531 & 0.3826 & 0.2618 & 0.7293 & 0.4567 & 113.6 & 79.4 \\
 & Ours ($\calL_{\text{ASP}}$) & \textbf{0.5032} & 0.4308 & \textbf{0.3670} & \textbf{0.2235} & \textbf{0.7210} & \textbf{0.4356} & 65.1 & \textbf{9.9} \\ \hline
\multirow{5}{*}{$\text{MIST}_\text{HER2}$} & CycleGAN & 0.1914 & 0.5633 & 0.6346 & 0.4695 & 0.8871 & 0.6386 & 240.3 & 311.1 \\
 & CUT+$\calL_{\text{GP}}$ & 0.1810 & 0.5321 & 0.4826 & 0.3060 & 0.8323 & 0.5383 & 66.8 & 19.0 \\
 & Pix2Pix & 0.1559 & 0.5516 & 0.5070 & 0.3253 & 0.8511 & 0.5588 & 137.3 & 82.9 \\
 & PyramidP2P & \textbf{0.2078} & 0.4787 & 0.4524 & 0.3313 & 0.8423 & 0.5262 & 104.0 & 61.8 \\
 & Ours ($\calL_{\text{ASP}}$) & 0.2004 & \textbf{0.4534} & \textbf{0.4150} & \textbf{0.2665} & \textbf{0.8174} & \textbf{0.4881} & \textbf{51.4} & \textbf{12.4} \\ \hline
\multirow{5}{*}{$\text{MIST}_\text{ER}$} & CycleGAN & 0.1982 & 0.5175 & 0.5092 & 0.3710 & 0.8672 & 0.5662 & 125.7 & 95.1 \\
 & CUT+$\calL_{\text{GP}}$ & \textbf{0.2217} & 0.4531 & 0.4079 & 0.2725 & \textbf{0.8194} & 0.4882 & 43.7 & 8.7 \\
 & Pix2Pix & 0.1500 & 0.5818 & 0.5282 & 0.3700 & 0.8620 & 0.5855 & 128.1 & 79.0 \\
 & PyramidP2P & 0.2172 & 0.4767 & 0.4538 & 0.3757 & 0.8567 & 0.5407 & 107.4 & 84.2 \\
 & Ours ($\calL_{\text{ASP}}$) & 0.2061 & \textbf{0.4336} & \textbf{0.4007} & \textbf{0.2649} & 0.8205 & \textbf{0.4799} & \textbf{41.4} & \textbf{5.8} \\ \hline
  \multicolumn{10}{l}{Note that KID values multiplied by 1000 are shown. CUT is from \cite{park2020contrastive}.}
\end{tabular}
\end{table}

\begin{table}[t]
\caption{
Ablation studies comparing different adaptive strategies for the SP loss.
The \textbf{best} and \uline{second best} values are highlighted.
Note that $\calL_{\text{SP}}$ is $\calL_{\text{ASP}}^{\text{(zero,uniform)}}$.
}
\label{tbl:ablation}
\centering
\begin{tabular}{clcccccccc}
\hline
\multirow{2}{*}{Dataset} & \multicolumn{1}{c}{\multirow{2}{*}{Method}} & \multirow{2}{*}{SSIM$\uparrow$} & \multicolumn{5}{c}{$\text{PHV}_{T=0.01}$$\downarrow$} & \multirow{2}{*}{FID$\downarrow$} & \multirow{2}{*}{KID$\downarrow$} \\
 & \multicolumn{1}{c}{} &  & layer1 & layer2 & layer3 & layer4 & avg. &  &  \\ \hline
\multirow{5}{*}{$\text{BCI}_\text{HER2}$} & $\calL_{\text{SP}}$ & 0.5094 & 0.4413 & \uline{0.3771} & \uline{0.2273} & \textbf{0.7202} & \uline{0.4415} & 62.7 & 11.4 \\
 & $\calL_{\text{ASP}}^{\text{(lambda,top)}}$ & \uline{0.5206} & 0.4411 & 0.3813 & 0.2282 & 0.7244 & 0.4438 & \uline{62.1} & 12.8 \\
 & $\calL_{\text{ASP}}^{\text{(lambda,lin.)}}$ & 0.5032 & \textbf{0.4308} & \textbf{0.3670} & \textbf{0.2235} & \uline{0.7210} & \textbf{0.4356} & 65.1 & \uline{9.9} \\
 & $\calL_{\text{ASP}}^{\text{(sigmoid,top)}}$ & \textbf{0.5236} & 0.4503 & 0.3877 & 0.2331 & 0.7292 & 0.4501 & 65.9 & 12.3 \\
 & $\calL_{\text{ASP}}^{\text{(linear,top)}}$ & 0.4890 & \uline{0.4327} & 0.3828 & 0.2305 & 0.7318 & 0.4445 & \textbf{61.9} & \textbf{9.8} \\ \hline
\multirow{5}{*}{$\text{MIST}_\text{HER2}$} & $\calL_{\text{SP}}$ & \textbf{0.2159} & 0.4712 & 0.4243 & 0.2611 & \uline{0.8129} & 0.4924 & 55.6 & 20.8 \\
 & $\calL_{\text{ASP}}^{\text{(lambda,top)}}$ & 0.2035 & \textbf{0.4451} & \textbf{0.4068} & \textbf{0.2554} & \textbf{0.8117} & \textbf{0.4798} & \uline{51.2} & 16.7 \\
 & $\calL_{\text{ASP}}^{\text{(lambda,lin.)}}$ & 0.2004 & \uline{0.4534} & \uline{0.4150} & 0.2665 & 0.8174 & \uline{0.4881} & 51.5 & \uline{12.4} \\
 & $\calL_{\text{ASP}}^{\text{(sigmoid,top)}}$ & \uline{0.2086} & 0.4655 & 0.4191 & \uline{0.2581} & 0.8138 & 0.4891 & \textbf{45.2} & \textbf{11.5} \\
 & $\calL_{\text{ASP}}^{\text{(linear,top)}}$ & 0.1809 & 0.4766 & 0.4262 & 0.2667 & 0.8178 & 0.4968 & 68.8 & 28.9 \\ \hline
\multirow{5}{*}{$\text{MIST}_\text{ER}$} & $\calL_{\text{SP}}$ & \textbf{0.2236} & 0.4517 & 0.4117 & 0.2714 & \uline{0.8208} & 0.4889 & 46.4 & 12.5 \\
 & $\calL_{\text{ASP}}^{\text{(lambda,top)}}$ & 0.2096 & 0.4388 & 0.4052 & \uline{0.2676} & 0.8215 & 0.4833 & \uline{42.4} & \uline{8.1} \\
 & $\calL_{\text{ASP}}^{\text{(lambda,lin.)}}$ & 0.2061 & \textbf{0.4336} & \uline{0.4007} & \textbf{0.2649} & \textbf{0.8205} & \textbf{0.4799} & \textbf{41.4} & \textbf{5.8} \\
 & $\calL_{\text{ASP}}^{\text{(sigmoid,top)}}$ & \uline{0.2192} & \uline{0.4376} & \textbf{0.3965} & 0.2684 & 0.8215 & \uline{0.4810} & 43.6 & \uline{8.1} \\
 & $\calL_{\text{ASP}}^{\text{(linear,top)}}$ & 0.1981 & 0.4581 & 0.4072 & 0.2706 & 0.8217 & 0.4894 & 46.9 & 10.7 \\ \hline
\end{tabular}
\end{table}

\noindent \textbf{Datasets.}
The following datasets are used in our experiments:
the Breast Cancer Immunohistochemical (BCI) challenge dataset \cite{liu2022bci} and our own MIST dataset that is now in the public domain.
The publicly available portion of BCI contains 3396 H\&E-HER2 patches for training and 500 of the same for testing.
Note that we have additionally normalized the brightness levels of all BCI images to the same level.
Due to the page limit, from the MIST dataset, here we only present detailed results on HER2 and ER.
For $\text{MIST}_\text{HER2}$, we extracted 4642 paired patches for training and 1000 for testing from 64 WSIs.
And for $\text{MIST}_\text{ER}$, we extracted 4153 patches for training, and 1000 for testing from 56 WSIs.
All patches are of size $1024\times1024$ and non-overlapping.
\textit{Additional results on $\text{MIST}_\text{Ki67}$ and $\text{MIST}_\text{PR}$ are provided in the Supplementary Materials.}

\noindent\textbf{Implementation Details.}
For all of our models, we used ResNet-6Blocks as the generator and a 5-layer PatchGAN as the discriminator.
We trained our networks with random $512\times512$ crops and a batch size of one.
The Adam optimizer \cite{kingma2014adam} was used with a linear decay scheduler (as shown in \cref{fig:functions}(c)) and an initial learning rate of $2\times10^{-4}$. 
The hyperparameters in \cref{eq:overall} are set as: $\lambda_{\text{PatchNCE}} = 10.0$, $\lambda_{\text{ASP}} = 10.0$ and $\lambda_{\text{GP}} = 10.0$.

\noindent\textbf{Evaluation Metrics.}
We compare the methods using both paired and unpaired evaluation metrics.
To compare a pair of images, generated and groundtruth, we use the standard SSIM (Structural Similarity Index Measure) and PHV (Perceptual Hash Value) as described in \cite{liu2021unpaired}. 
As for the unpaired metrics, we use the FID (Fréchet Inception Distance) and the KID (Kernel Inception Distance).

\noindent\textbf{Qualitative Evaluations.}
In \cref{fig:qualitative}, we compare visually the generated IHC images by our framework.
It can be observed that by using either $\calL_{\text{SP}}$ or $\calL_{\text{ASP}}$, the pathological representations in the generated images are significantly more accurate.
And by using $\calL_{\text{ASP}}$, such representations appear to be more consistent.

\noindent\textbf{Quantitative Evaluations.}
The full results comparing existing I2IT methods to ours are tabulated in \cref{tbl:competing}.
Overall, it can be observed that the proposed framework with the ASP loss consistently outperforms existing methods across all three datasets.
Subsequently, in \cref{tbl:ablation}, we further provide results using different weighting and scheduling functions in our proposed $\calL_{\text{ASP}}$.
With $\calL_{\text{SP}}$ already being a strong baseline, using different adaptive strategies can provide further gains in performance.
It is also worth noting that if reinforced prematurely, adaptive weighting can lead to inferior convergence, \eg $\calL_{\text{ASP}}^{\text{(linear,top)}}$.

\section{Conclusion}
In this paper, we have proposed the Adaptive Supervised PatchNCE (ASP) loss for learning H\&E-to-IHC stain translation with inconsistent GT image pairs.
The adaptive logic in ASP is based on the intuition that inconsistent patch locations should contribute less to learning.
We demonstrated that our proposed framework is able to achieve significant improvements both qualitatively and quantitatively over the existing approaches for translations to multiple IHC stains.
Finally, we are making public our Multi-IHC Stain Translation dataset with the hope to assist further research towards accurate H\&E-to-IHC stain translation.

\bibliographystyle{splncs04}
\bibliography{egbib}

\newpage
\section{Supplementary Materials}
\begin{table}[h]
\caption{
Additional quantitative evaluations on $\text{MIST}_\text{Ki67}$ and $\text{MIST}_\text{PR}$. 
Ours is with $\calL_{\text{ASP}}^{\text{(lambda,linear)}}$.
}
\centering
\begin{tabular}{clcccccccc}
\hline
\multirow{2}{*}{Dataset} & \multicolumn{1}{c}{\multirow{2}{*}{Method}} & \multirow{2}{*}{SSIM$\uparrow$} & \multicolumn{5}{c}{$\text{PHV}_{T=0.01}$$\downarrow$} & \multirow{2}{*}{FID$\downarrow$} & \multirow{2}{*}{KID$\downarrow$} \\
 & \multicolumn{1}{c}{} &  & layer1 & layer2 & layer3 & layer4 & avg. &  &  \\ \hline
\multirow{5}{*}{$\text{MIST}_\text{Ki67}$} & CycleGAN & \textbf{0.3875} & 0.8274 & 0.8275 & 0.6081 & 0.9038 & 0.7917 & 343.9 & 317.9 \\
 & CUT+$\calL_{\text{GP}}$ & 0.1909 & 0.5426 & 0.4739 & 0.3160 & 0.8415 & 0.5435 & 76.1 & 43.5 \\
 & Pix2Pix & 0.1819 & 0.5468 & 0.4905 & 0.3415 & 0.8496 & 0.5571 & 147.0 & 142.4 \\
 & PyramidP2P & 0.2286 & 0.4533 & 0.4222 & 0.3360 & 0.8363 & 0.5120 & 94.4 & 78.0 \\
 & Ours ($\calL_{\text{ASP}}$) & 0.2410 & \textbf{0.4472} & \textbf{0.4001} & \textbf{0.2701} & \textbf{0.8128} & \textbf{0.4826} & \textbf{51.0} & \textbf{19.1} \\ \hline
\multirow{5}{*}{$\text{MIST}_\text{PR}$} & CycleGAN & 0.2232 & 0.5334 & 0.5554 & 0.3867 & 0.8654 & 0.5852 & 96.1 & 96.6 \\
 & CUT+$\calL_{\text{GP}}$ & 0.2153 & 0.4656 & 0.4128 & 0.2724 & 0.8154 & 0.4916 & 54.6 & 20.1 \\
 & Pix2Pix & 0.1617 & 0.6027 & 0.5569 & 0.4043 & 0.8601 & 0.6060 & 183.8 & 148.1 \\
 & PyramidP2P & \textbf{0.2403} & 0.5078 & 0.4682 & 0.3509 & 0.8446 & 0.5429 & 98.8 & 59.5 \\
 & Ours ($\calL_{\text{ASP}}$) & 0.2159 & \textbf{0.4484} & \textbf{0.3898} & \textbf{0.2564} & \textbf{0.8080} & \textbf{0.4757} & \textbf{44.8} & \textbf{10.2} \\ \hline
 \multicolumn{10}{l}{Note that KID values multiplied by 1000 are shown.}
\end{tabular}
\end{table}

\begin{table}[h!]
\caption{
Additional ablation studies on $\text{MIST}_\text{Ki67}$ and $\text{MIST}_\text{PR}$.
}
\centering
\begin{tabular}{clcccccccc}
\hline
\multirow{2}{*}{Dataset} & \multicolumn{1}{c}{\multirow{2}{*}{Method}} & \multirow{2}{*}{SSIM$\uparrow$} & \multicolumn{5}{c}{$\text{PHV}_{T=0.01}$$\downarrow$} & \multirow{2}{*}{FID$\downarrow$} & \multirow{2}{*}{KID$\downarrow$} \\
 & \multicolumn{1}{c}{} &  & layer1 & layer2 & layer3 & layer4 & avg. &  &  \\ \hline
\multirow{5}{*}{$\text{MIST}_\text{Ki67}$} & $\calL_{\text{SP}}$ & 0.2089 & 0.4631 & 0.4014 & 0.2686 & 0.8200 & 0.4883 & \textbf{41.3} & \textbf{8.3} \\
 & $\calL_{\text{ASP}}^{\text{(lambda,top)}}$ & 0.2078 & 0.4507 & 0.3938 & \uline{0.2642} & \textbf{0.8103} & \uline{0.4798} & 50.4 & 21.5 \\
 & $\calL_{\text{ASP}}^{\text{(lambda,linear)}}$ & \textbf{0.2410} & \uline{0.4472} & \uline{0.4001} & 0.2701 & \uline{0.8128} & 0.4826 & 51.0 & 19.1 \\
 & $\calL_{\text{ASP}}^{\text{(sigmoid,top)}}$ & \uline{0.2116} & \textbf{0.4403} & \textbf{0.3940} & \textbf{0.2634} & 0.8132 & \textbf{0.4777} & \uline{41.4} & \uline{11.0} \\
 & $\calL_{\text{ASP}}^{\text{(linear,top)}}$ & 0.1819 & 0.4483 & 0.4067 & 0.2762 & 0.8154 & 0.4867 & 49.5 & 21.4 \\ \hline
\multirow{5}{*}{$\text{MIST}_\text{PR}$} & $\calL_{\text{SP}}$ & \textbf{0.2243} & 0.4511 & 0.3996 & 0.2682 & 0.8163 & 0.4838 & 50.7 & 11.2 \\
 & $\calL_{\text{ASP}}^{\text{(lambda,top)}}$ & \uline{0.2169} & \uline{0.4495} & \uline{0.3940} & \uline{0.2604} & \uline{0.8130} & \uline{0.4792} & \textbf{43.6} & 11.4 \\
 & $\calL_{\text{ASP}}^{\text{(lambda,linear)}}$ & 0.2159 & \textbf{0.4484} & \textbf{0.3898} & \textbf{0.2564} & \textbf{0.8080} & \textbf{0.4757} & \uline{44.8} & \textbf{10.2} \\
 & $\calL_{\text{ASP}}^{\text{(sigmoid,top)}}$ & 0.2056 & 0.4725 & 0.4105 & 0.2703 & 0.8150 & 0.4921 & 48.6 & 11.6 \\
 & $\calL_{\text{ASP}}^{\text{(linear,top)}}$ & 0.2164 & 0.4569 & 0.4017 & 0.2655 & \uline{0.8130} & 0.4843 & 45.5 & \uline{10.9} \\ \hline
\end{tabular}
\end{table}

\begin{figure}[h]
\centering
\includegraphics[width=1.0\textwidth]{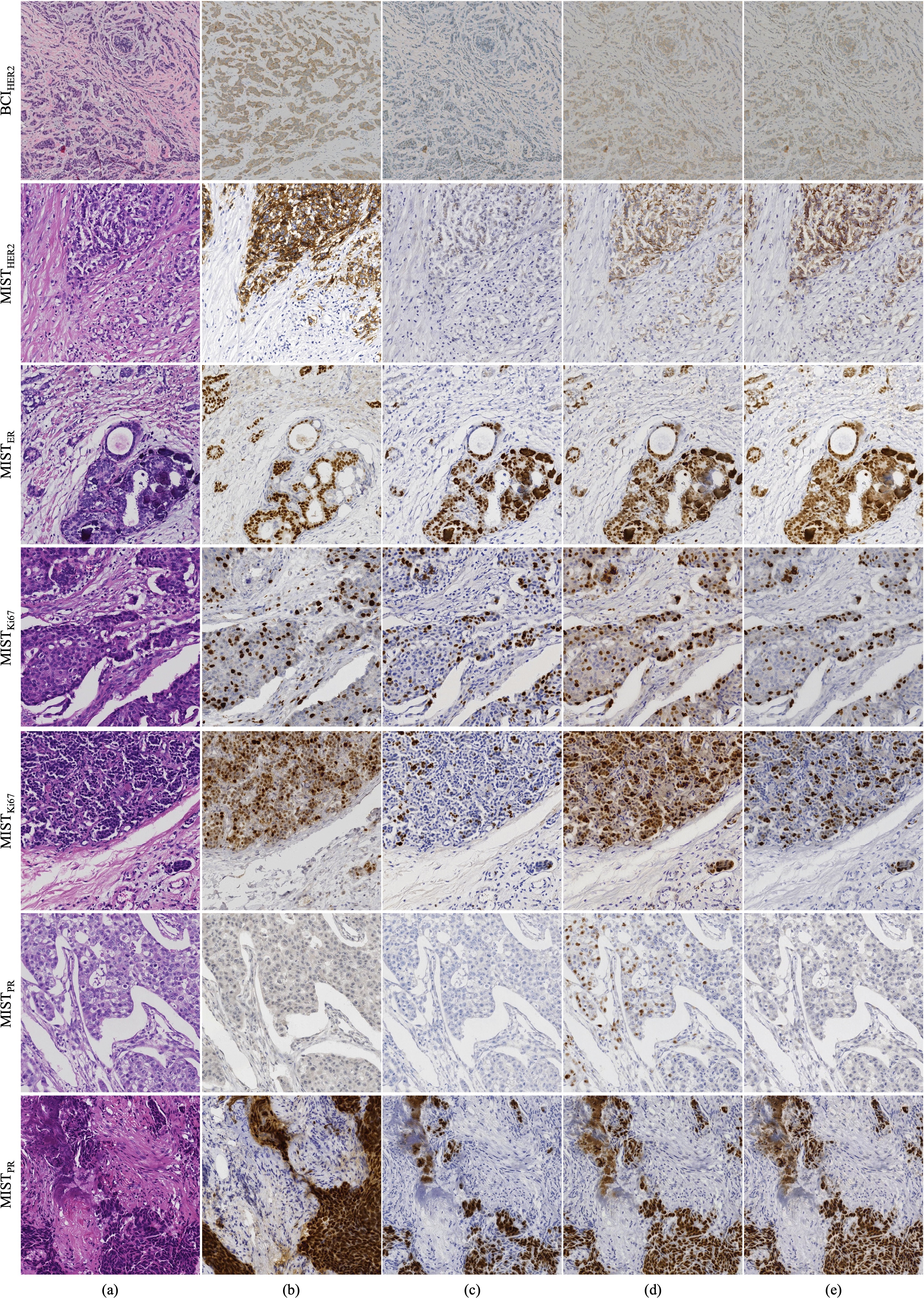}
\caption{
Left to right: (a) Input H\&E image; (b) Groundtruth IHC image; (c) Generated image without $\calL_{\text{SP}}$; (d) With $\calL_{\text{SP}}$; (e) With $\calL_{\text{ASP}}^{\text{(lambda,linear)}}$.}
\label{fig:qualitative}
\end{figure}

\end{document}